\pdfoutput=1

\documentclass[11pt]{article}

\usepackage[]{acl}
\usepackage{graphicx}
\usepackage{multirow}
\usepackage{times}
\usepackage{latexsym}
\usepackage{amsfonts,amssymb}
\usepackage[T1]{fontenc}

\usepackage[utf8]{inputenc}
\usepackage{amsmath}
\usepackage{microtype}

\title{Focus-Driven Contrastive Learning for Medical Question Summarization}

\author{Ming Zhang$^1$$^,$$^2$ \ \ Shuai Dou$^3$ \ \  Ziyang Wang$^1$$^,$$^2$ \ \ Yunfang Wu$^1$$^,$$^3$\thanks{*Corresponding author.}\\
   $^1$MOE, Key Laboratory of Computational Linguistics, Peking University\\
   $^2$School of Software and Microelectronics, Peking University \\
   $^3$School of Computer Science, Peking University\\
  $^1${\tt \{zhangming,wzy232303\}@stu.pku.edu.cn}\\  
  $^2${\tt \{monkdou,wuyf\}@pku.edu.cn}
}

\begin{document}
\maketitle
\begin{abstract}
Automatic medical question summarization can significantly help the system to understand consumer health questions and retrieve correct answers.
The Seq2Seq model based on maximum likelihood estimation (MLE) has been applied in this task, 
which faces two general problems: the model can not capture well question focus and and the traditional MLE strategy lacks the ability to understand sentence-level semantics.  
To alleviate these problems, we propose a novel question focus-driven contrastive learning framework (QFCL).
Specially, we propose an easy and effective approach to generate hard negative samples based on the question focus, and exploit contrastive learning at both encoder and decoder to obtain better sentence-level representations. 
On three medical benchmark datasets, our proposed model achieves new state-of-the-art results, and obtains a performance gain of 5.33, 12.85 and 3.81 points over the baseline BART model on three datasets respectively. Further human judgement and detailed analysis prove that our QFCL model learns better sentence representations with the ability to distinguish different sentence meanings, and generates high-quality summaries by capturing question focus.  
\end{abstract}

\section{Introduction}
\begin{table}[ht]
\resizebox{\linewidth}{!}{%
\begin{tabular}{|l|}
\hline
\textbf{Input question: consumer health question (CHQ)}\\
subject: \textcolor[rgb]{0,0.6902,0.3137}{gender dysphoria} message: no health care on\\
my son suffering from \textcolor[rgb]{0,0.6902,0.3137}{gender dysphoria} what can we\\
do to help him he worked out of high school no problems\\
now not working and about shutting himself in his room \\
24/7 theres nothing this condition in our area we live in\\ 
\textsf{[}location\textsf{]}.no help in area what can we do he has had bad\\
thoughts already please help us with some sort of info\\
thank yuo \textsf{[}name\textsf{]} \textsf{[}location\textsf{]} \\
\hline
\textbf{Golden summary: frequently asked question (FAQ):}\\
Where can I find information on treatment and resources\\
for \textcolor[rgb]{0,0.6902,0.3137}{gender dysphoria}?\\
\hline
\hline
\textbf{Summary by BART (baseline):} \\
What are the treatments for \textcolor[rgb]{1,0,0}{weight loss}? \\
\hline
\textbf{Summary by our model:} \\ 
What are the treatments for \textcolor[rgb]{0,0.6902,0.3137}{gender dysphoria}? \\
\hline
\end{tabular}
}
\caption{An example of medical question summarization in MeqSum dataset, where the question focus is highlighted in green. 
Summaries generated by BART and our model are also listed.}
\label{fig1}
\end{table}

A growing number of health questions are raised by consumers on websites nowadays, which are usually written in natural language and including detailed and peripheral information not related to the answers. Summaries of such questions can greatly improve the performance in retrieving relevant answers \cite{ben-abacha-demner-fushman-2019-summarization}. Accordingly, the medical question summarization task is defined as summarizing the consumer health questions (CHQ) into frequently asked questions (FAQ), which are shorter but remain essential information of the original question to get correct answers. An example of medical question summarization is shown in Table \ref{fig1}.

The Seq2Seq neural models have been widely used in abstractive summarization  \cite{nallapati-etal-2016-abstractive,lewis-etal-2020-bart,zhang2020pegasus} and show promising potentials, 
and they have also been applied in medical question summarization and achieve current state-of-the-art results. 
\citet{ben-abacha-demner-fushman-2019-summarization} apply the pointer-generator model for this task. \citet{yadav-etal-2021-reinforcement} present a reinforcement learning framework with question-type identification reward and question-focus recognition reward. \citet{mrini-etal-2021-gradually} propose a multitask learning method by treating recognizing question entailment as an auxiliary task. 

\begin{figure}[h]
\centering
\small
\includegraphics[width=\linewidth]{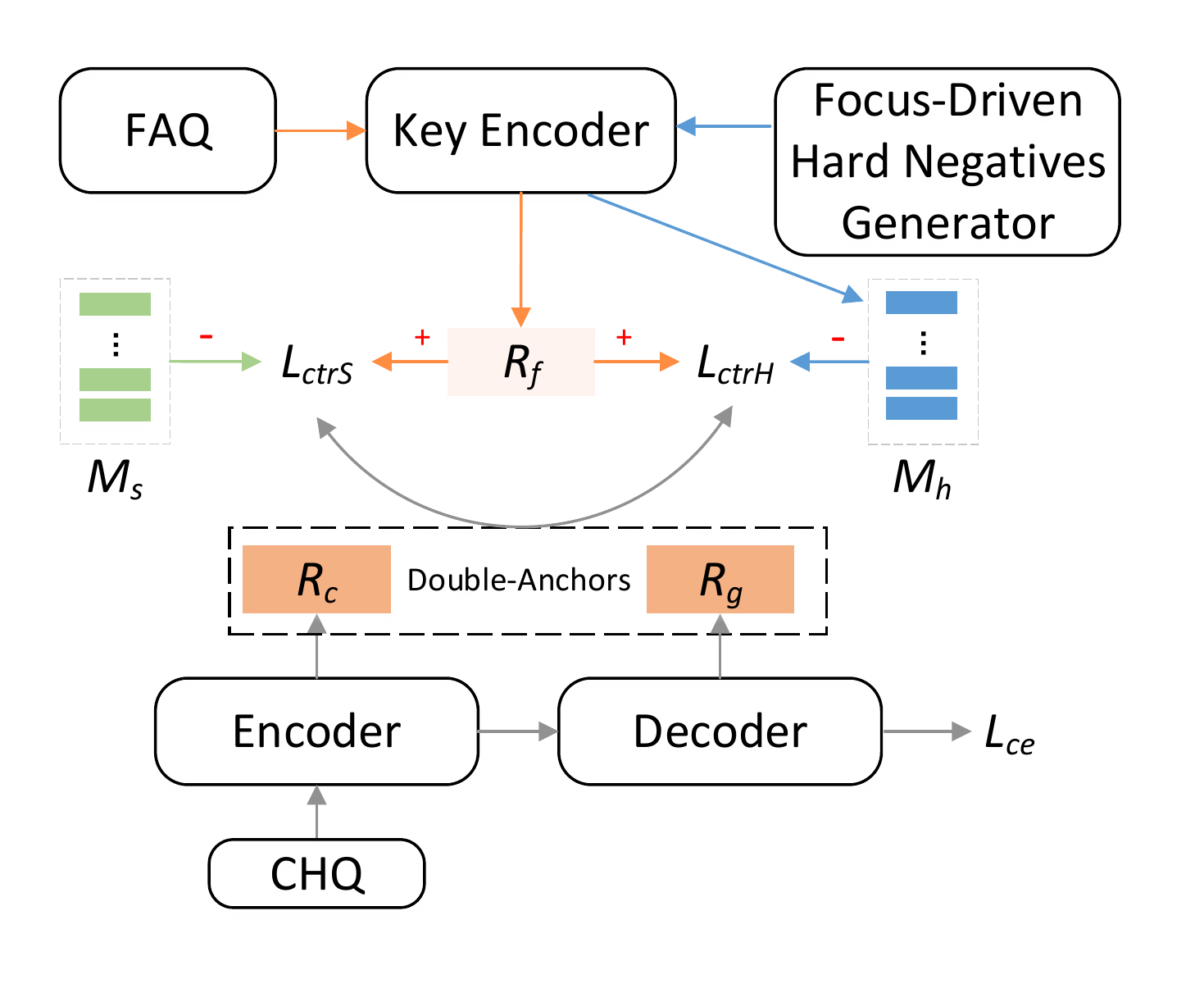} 
\caption{Sketch of our proposed contrastive learning framework. $M_s$,$M_h$ represents the memory bank that contains simple negative samples and hard negative samples respectively. $\mathcal{R}_f$, $\mathcal{R}_c$, $\mathcal{R}_g$ denotes the sentence representation of FAQ, CHQ and generated summary. $\mathcal{L}_{ctrS}$ and $\mathcal{L}_{ctrH}$ are contrastive learning loss on simple negative samples and hard negative samples respectively. $+$ indicates the positive sample, and $-$ indicates the negative sample.}.
\label{pipeline-sketch}
\end{figure}

In the medical question summarization task, the input question CHQ is always lengthy and contains redundant information, where some salient medical entities and the semantic focus of question are vital to understand users' intention. But it still remains a challenging task for the existing methods to capture the question focus. As described in the example \ref{fig1}, the focus "\textit{gender dysphoria}" is mis-replaced by "\textit{weight loss}" in the summary generated by the fine-tuned BART, resulting in a completely different meaning from the original sentence.


For the medical question summarization task, the generated question summary is required to semantically close to the reference question. However, in most of current pre-trained models such as BART \cite{lewis-etal-2020-bart}, the model adopts  maximum likelihood estimation (MLE) and mainly focuses on the accuracy of the prediction of masked tokens, but does not guarantee to the semantic similarity or dissimilarity of the whole sentences. 
To address this issue, 
some previous works adopt reinforcement learning (RL) in text summarization task \cite{li-etal-2019-deep, paulus2018a}, but RL suffers from the noise gradient estimation problem \cite{greensmith2004variance}, which makes the training process unstable and sensitive to hyper-parameters.

To alleviate these problems, we propose a novel question focus-driven contrastive learning (QFCL) framework for medical question summarization, as illustrated in Figure \ref{pipeline-sketch}.  In our model, we introduce a "\textit{double anchors}" strategy for contrastive learning, by utilizing the sentence representation of CHQ as an anchor and the generated summary as another anchor, and regarding the golden reference FAQ as the positive sample. 
In addition, we present a "\textit{focus-driven hard negatives generator}" to construct hard negative samples, by replacing the focus phrases with other phrases sharing the same attribute. 

Through contrastive learning, we minimize the distance between CHQ/generated summary and golden reference,  
and maximize the distance between CHQ/generated summary and other negative samples. 
By using the \textit{double anchors}, our model is able to extract sentence-level semantic features to alleviate the problem of MLE. With the help of \textit{hard negatives generator}, the model learns to pay more attention to question focus and thus produces high quality summary.   

We conduct extensive experiments on three medical question summarization datasets: Meqsum \cite{ben-abacha-demner-fushman-2019-summarization}, HealthCareMagic and iCliniq  \cite{zeng-etal-2020-meddialog}. Our proposed model outperforms previous best results by a wide margin, achieving new state-of-the-art results on all three datasets. Compared with the baseline BART, our model brings a relative performance gain of $12.2\%$, $28.7\%$ and $9.6\%$ on Meqsum, Cliniq and HealthcareMagic respectively. 
Through analysis, we prove that our model significantly gains the power of distinguishing the semantics between generated summaries and negative samples, and our model generates high-quality summaries capturing more question focuses. Our data and codes are publicly available for future research \footnote{https://github.com/zhangming880102/QFCL}.

\begin{figure*}[ht]
\centering
\includegraphics[width=0.8\textwidth]{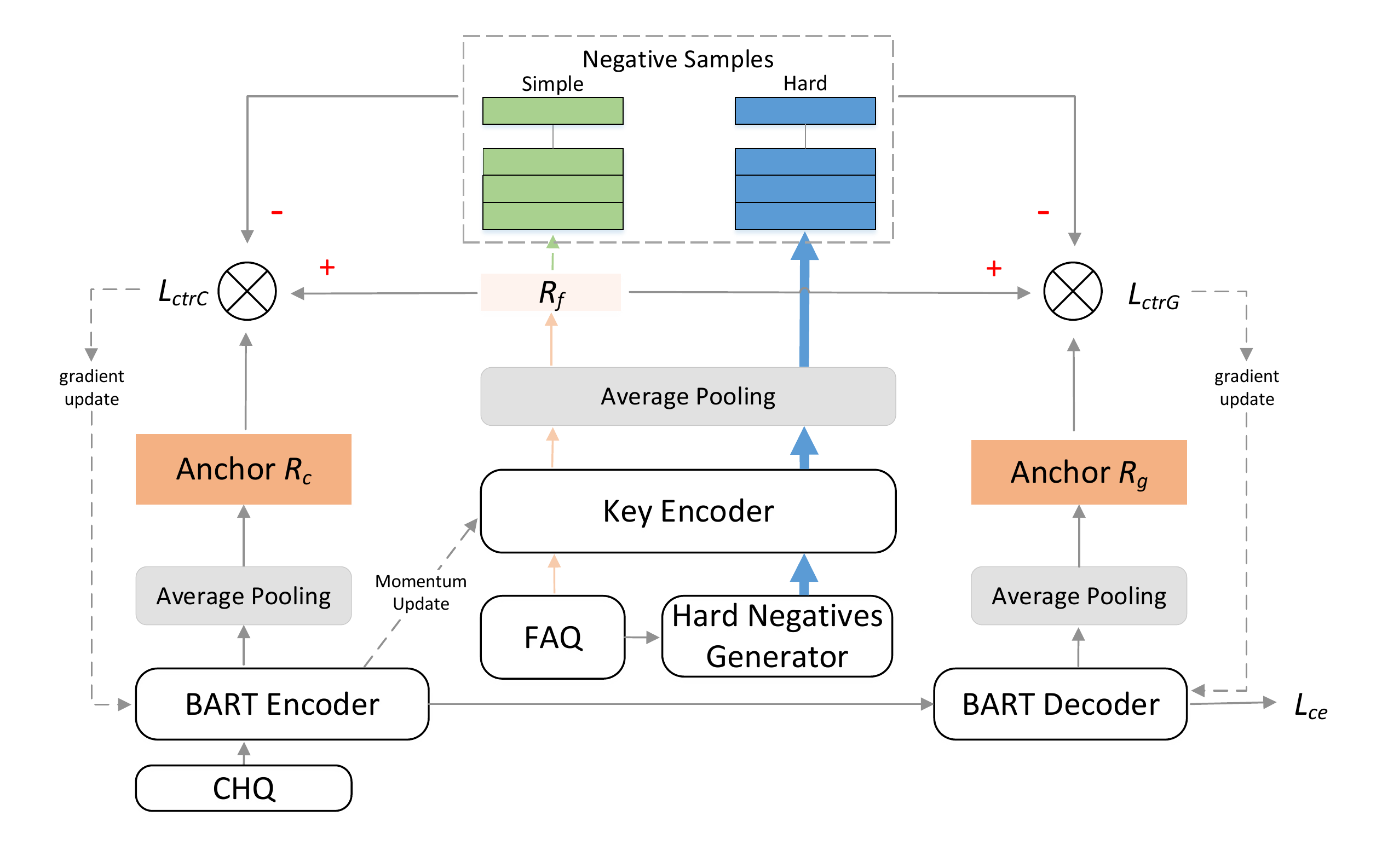} 
\caption{The overall framework of QFCL. $\mathcal{L}_{ctrC}$ and $\mathcal{L}_{ctrG}$ are contrastive learning loss on the two anchors respectively. }
\label{framework}
\end{figure*}
\section{Ralated Work}
\subsection{Medical Question Summarization}

The medical question summarization task is defined by \citet{ben-abacha-demner-fushman-2019-summarization}. They construct a benchmark dataset Meqsum, and apply a pointer-generator model to generate question summary. 
At the question summarization campaign of MEDIQA-21 organized by \citet{ben-abacha-etal-2021-overview}, almost all approaches rely on the fine-tuning of pre-trained transformer models. Transfer learning, knowledge-base, and ensemble methods are widely utilized by participanting teams to achieve better performance \cite{he-etal-2021-damo,yadav-etal-2021-nlm,mrini-etal-2021-ucsd,sanger-etal-2021-wbi}. In this paper, we also base our method on the strong pre-trained BART model.

Recently, \citet{yadav-etal-2021-reinforcement} propose a RL framework with two question-aware semantic rewards: question-type identification reward (QTR) and question-focus recognition reward (QFR). QTR is to identify whether the question types are consistent with the gold question, and QFR is designed to capture question focus. 
But in their work, the question types and question focuses in the dataset should be manually labeled, which is both time-consuming and labor-intensive for large-scale datasets such as HealthcareMagic and iCliniq. Moreover, the RL training process is unstable.
\citet{mrini-etal-2021-gradually} 
claim an equivalence between medical question summary and recognizing question entailment(RQE), 
and employ multi-task learning to train the model to not only perform next-word-prediction but also carry question entailment recognition. These two studies demonstrate that the pre-trained models achieve better performance after capturing the underlying sentence semantics of generated questions. Different from these works, we exploit contrastive learning to obtain 
focus-aware question representations.


\subsection{Contrastive Learning}
Different from the traditional methods which learn representations in pixel-level for computer vision tasks, contrastive learning encodes high-level features to distinguish different objects and has achieved great success \cite{henaff2020data,chen2020simple,9156540,he2020momentum}, and it has also been applied in several NLP tasks such as machine translation \cite{pan-etal-2021-contrastive}, pre-training \cite{chi-etal-2021-infoxlm} and question answering \cite{yang-etal-2021-xmoco}. In the field of summarization, \citet{liu-liu-2021-simcls} present a contrastive framework to bridge the gap between the learning objective and evaluation metrics, \citet{cao-wang-2021-cliff} design several negative sample construction strategies to solve the factual inconsistency problem. In contrast, we use the MoCo structure to handle with the large volume of negative samples, and propose a new negative sample construction method.

\citet{chen2020simple} prove that large size of negative samples can improve the performance of contrastive learning, but it also brings heavy burden on computation cost. To address this issue, \citet{he2020momentum} propose MoCo, which maintains a queue as the memory bank to store negative samples. 
MoCo adopts two encoders with the same structure: key encoder and query encoder, where the key encoder is momentum updated from the query encoder.

\section{Model}
Given an input question CHQ, which is written by consumers and contains lengthy and complex information, the medical question summarization task aims to automatically generate a question summary that is a frequently asked question (FAQ), capturing the essential information to help efficiently retrieve correct answers. A more detailed structure of our proposed QFCL model is presented  
in Figure \ref{framework}.


\subsection{Contrastive Learning Architecture}
We employ the pre-trained BART \cite{lewis-etal-2020-bart} as our basic model to generate question summaries. For contrastive learning, we adopt the MoCo architecure \cite{he2020momentum}, which contains a key encoder $E_k$  with the same structure as the BART encoder $E_q$, and a queue to store simple negative samples with large volume. The simple negative samples in the queue are progressively replaced by current mini-batch of representations extracted from the key encoder. All samples in the queue will be used as negative samples in the next batch. In addition, QFCL employs a \textit{hard negatives generator} to generate hard negative samples.

In our model, the BART encoder $E_q$ and the decoder are updated via back propagation by combining three types of loss functions, as described in the subsequent sections. The parameters of $E_k$ are frozen and updated slowly towards that of $E_q$: 

\begin{equation}
\theta_{k} \leftarrow m\theta_{k} + (1-m)\theta_{q}
\label{eq1}
\end{equation}
where $m$ is a momentum coefficient.

At the inference, only the BART encoder and decoder are retained,  other parts such as the key encoder, the queue, and the \textit{hard negatives generator} are all discarded.

\subsection{Simple Negative Samples}
In the medical question summarization task, the input question CHQ should be semantically close to its reference summary FAQ but different from other question summaries. Therefore,
we regard the CHQ $c_i$ in the $i$-th pair as the anchor, FAQ $f_i$ in the same pair as the positive sample and randomly select $f_j$ from other different pairs to serve as simple negative samples. 

Let $\mathcal{R}_{s}$ denote the average decoded output of an arbitrary sentence $s$, the objective function of the simple contrastive learning is defined as:

\begin{equation}
\mathcal{L}_{ctrCS} = -log{\frac{e^{{sim(\mathcal{R}_{ci},\mathcal{R}_{fi})}/{\tau}}}{\sum_{\mathcal{R}_{fj} \in M_s}{e^{{sim(\mathcal{R}_{ci},\mathcal{R}_{fj})}/{\tau}}}}}
\label{eq2}
\end{equation}
where $\mathcal{R}_{ci}$ indicates the sentence representation of the $i$-th CHQ extracted from $E_q$, and $\mathcal{R}_{fi}$ and $\mathcal{R}_{fj}$ are extracted from the key encoder $E_k$ for the $i$-th and $j$-th FAQ respectively. The operation $sim$ is to calculate the cosine similarity, $\tau$ is a temperature hyper-parameter. $M_s$ is the memory bank which contains one positive sample and $K$ simple negative samples in the queue with respect to an anchor.



\subsection{Focus-Driven Hard Negative Samples}\label{section:hard-gen}
\begin{figure}[h]
\centering
\small
\includegraphics[width=\linewidth]{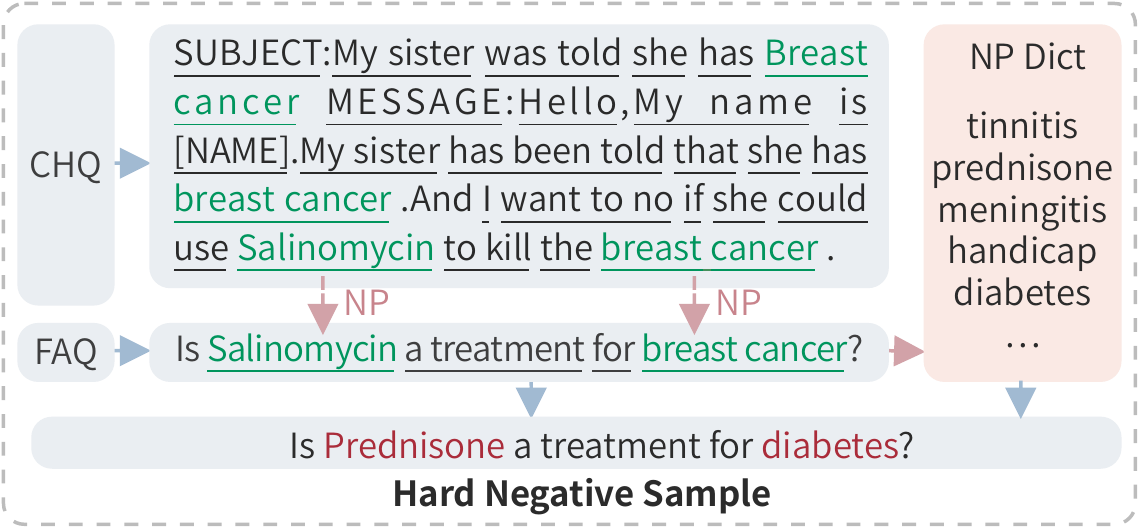} 
\caption{The method of hard negative samples generation.}
\label{hard-generator}
\end{figure}
The above simple negative samples are randomly selected. As claimed by \cite{NEURIPS2020_f7cade80}, hard negative samples that are more similar to positive samples can facilitate the model to get better performance. Inspired by this, we build a bridge between hard sample generation and question focus prediction. 

\subsubsection{Question Focus Identification}
As mentioned before, the question focus is essential to understand a consumer health question. If some focus phrases are missing in the generated summary, the semantic will 
drift far away from the original user's intention. So we construct difficult negative samples based on the question focus to enhance contrastive learning. Specially,   
we replace the focus phrases with some other phrases of the same attribution, and keep other words of the sentence unchanged. An example of hard negative sample generation is shown in Figure \ref{hard-generator}. 

One issue for our method is how to automatically annotate question focus. \citet{yadav-etal-2021-reinforcement} manually labeled the question focus in MeqSum dataset. However, this is quite time-consuming and labor-intensive, driving us to find a method which can automatically mark the question focus in larger datasets, such as HealthcareMagic and iCliniq. We analyzed the manually labeled MeqSum dataset, and found that in 340 of the total 500 records (up to $68\%$), the question focuses are the overlap phrases between CHQ and FAQ. 
Accordingly, we hypothesize that the same phrases appearing both in the source question and the golden summary have a high probability to be key-phrases. This idea is also proved to be effective in \cite{li2020keywords}.

Since the question focus is usually a phrase rather than a single word, we need to split one sentence into phrases. We apply the chunker \cite{akbik-etal-2018-contextual} to the CHQ and FAQ text, and record the chunk label of each phrase. Then the consistent phrases appearing both in CHQ and FAQ are labeled as the question focuses. 

\subsubsection{Hard Negative Sample Generation}
We constructed a dictionary by concatenating all phrases of the FAQ sentences in the train set. To generate hard negative samples, the question focuses are randomly replaced by other phrases of the same chunk label from the dictionary. As shown in Figure 2, “\textit{breast cancer}” is replaced by “\textit{diabetes}” since they share the same label “\textit{NP}”.  We repeat this process $N_h$  times to construct $N_h$  different hard negative samples for each CHQ-FAQ pair.

\subsubsection{Contrastive Learning on Hard Negative Samples}
The sentence representation of hard sample $\mathcal{R}_h$ is extracted from the key encoder $E_k$. We define the hard loss function of contrastive learning as:

\begin{equation}
\mathcal{L}_{ctrCH} = -log{\frac{e^{{sim(\mathcal{R}_{ci},\mathcal{R}_{fi})}/{\tau}}}{\sum_{\mathcal{R}_h \in M_h}{e^{{sim(\mathcal{R}_{ci},\mathcal{R}_h)}/{\tau}}}}}
\label{eq3}
\end{equation}
where $M_h$ denotes the memory bank containing one positive sample and $N_h$ hard negative samples.

This loss function forces the model to not only shorten the distance between CHQ and FAQ, but also expand the gap between the CHQ and hard negative samples. In this way, we achieve the goal of making the model pay more attention to the question focus, and obtain a focus-aware representation. 

\subsection{Contrastive Learning at Decoder}
An imbalance existing in the above method is that contrastive learning is only utilized at the encoder. 
We fine-tuned BART on iCliniq dataset, and found that the decoder lacks the ability to distinguish the representations 
between the generated summary and the positive samples/unrelated negative samples,
as \begin{small}$s_{g\_faq}^{+}$\end{small}, \begin{small}$s_{g\_sim}^{-}$\end{small}, \begin{small}$s_{g\_hard}^{-}$\end{small} shown in Figure \ref{similarity_and_epoch}. 
Therefore, we try to improve the similarity between the generated summary and its reference FAQ, and at the same time enlarge the dis-similarity between the generated summary and other unrelated questions. 

Specially, we regard the generated summary as an extra anchor, and denote the representation of the generated summary as $g_i$. Since the output summary should be semantically consistent with the corresponding FAQ, we consider the representation of the FAQ $f_i$ in the same pair as the positive sample, and select the simple negative samples randomly from the queue and generate hard negative samples using the \textit{hard negatives generator}. The object functions of contrast loss $\mathcal{L}_{ctrGS}$ and $\mathcal{L}_{ctrGH}$ at the decoder end are defined in a similar style as Equation \ref{eq2} and \ref{eq3}, except that the anchor $c_i$ is replaced by another anchor $g_i$.

\subsection{Overall Objective Function}
For predicting next tokens in the generated summary, we use the cross entropy loss $\mathcal{L}_{ce}$:

\begin{equation}
\mathcal{L}_{ce} = -\frac{1}{\left|T\right|}\sum_{t \in T}{log(p(y_t|x,y_{1:t-1},\theta))}
\label{eq4}
\end{equation}

In our model, the overall loss function consists of five parts: the cross entropy loss $\mathcal{L}_{ce}$ and four different loss functions of contrastive learning: $\mathcal{L}_{ctrCS}$, $\mathcal{L}_{ctrCH}$ for the anchor at the encoder end, $\mathcal{L}_{ctrGS}$, $\mathcal{L}_{ctrGH}$ for the anchor at the decoder end. We define the contrastive learning loss with respect to these two anchors as:

\begin{equation}
\begin{split}
\mathcal{L}_{ctrC} &=\alpha\mathcal{L}_{ctrCS} + \beta\mathcal{L}_{ctrCH} \\
\mathcal{L}_{ctrG} &=\alpha\mathcal{L}_{ctrGS} + \beta\mathcal{L}_{ctrGH} \\
\end{split}
\label{eq7}
\end{equation}
where $\alpha$, $\beta$ are hyper-parameters to control the balance between simple negatives and hard ones. The weights of contrastive learning loss at the encoder and decoder are considered as equal, and the overall loss is defined as:

\begin{equation}
\mathcal{L} = \mathcal{L}_{ce} + \frac{1}{2}\mathcal{L}_{ctrC}
+ \frac{1}{2}\mathcal{L}_{ctrG} 
\label{eq:overall}
\end{equation}

\section{Experiments}
\subsection{Datasets}
We conduct experiments on three English benchmark medical question summarization datasets, including Meqsum, HealthcareMagic and iCliniq. Meqsum is a high-quality dataset from NIH \footnote {\noindent www.nlm.nih.gov/medlineplus\label{nih_footnote}}, constructed by \citet{ben-abacha-demner-fushman-2019-summarization}. \citet{mrini-etal-2021-joint} extracted HealthCareMagic and iCliniq datasets from MedDialog \cite{zeng-etal-2020-meddialog} , which are collected automatically from the online healthcare service platforms \footnote {\noindent www.healthcaremagic.com\label{healthcare_footnote}} \footnote{\noindent www.icliniq.com\label{icliniq_footnotel}}. MeqSum's and HealthcareMagic's summaries are written by medical experts in formal style, while iCliniq's are patient-written. 
We list some statistics of these datasets in table \ref{tab:dataset}. Following previous works, we adopt ROUGE \cite{lin-2004-rouge}\footnote{\noindent https://pypi.org/project/py-rouge \label{rouge_footnote}}  as the evaluation metric. 
Our data and codes are available at https:// github.com/zhangming880102/QFCL.

\begin{table}[h]
\small
\centering
\resizebox{\linewidth}{!}{%
\begin{tabular}{l|r|r|r|r}
\hline
Dataset   &    Train   & Dev   & Test   & Length        \\ 
\hline
\hline
MeqSum                    & 400 & 100 &500 & 60.8/10.1 \\ 
HealthCareMagic & 181,122 & 22,641 & 22,642 & 82.8/9.7 \\ 
iCliniq                   & 24,851 & 3,105 & 3,106 & 89.7/12.3 \\ \hline
\end{tabular}%
}
\caption{Statistics of three medical question summarization datasets. Length indicates the average length of CHQ/FAQ.}
\label{tab:dataset}
\end{table}

\begin{table*}[ht]
\small
\centering
\resizebox{\textwidth}{!}{%
\begin{tabular}{p{20em}|ccc|ccc|ccc}
\hline
\multirow{2}{*}{\textbf{Model}} &\multicolumn{3}{c|}{\textbf{MeqSum}} &\multicolumn{3}{c|}{\textbf{iCliniq}} &\multicolumn{3}{c}{\textbf{HealthCareMagic}}\\ 
\cline{2-10} 
& \textbf{R1}   & \textbf{R2}            & \textbf{RL}  & \textbf{R1}   & \textbf{R2}            & \textbf{RL} & \textbf{R1}   & \textbf{R2}            & \textbf{RL}  \\ 
\hline
\hline
ProphetNet + QTR + QFR\cite{yadav-etal-2021-reinforcement} &   45.52 & 27.54 &48.19 & - & - & - & - & - & - \\
MTL+Data augmentation\cite{mrini-etal-2021-gradually}   & 49.20 & 29.50 &44.80 & 54.20 & 36.90 & 49.10 & 45.90 & 24.30 & 42.90 \\   \hline
\hline
BART \cite{lewis-etal-2020-bart} &   46.17 & 28.05 & 43.75 & 48.79 & 25.47 & 44.69 & 42.33  & 23.07 & 39.60 \\ 

BART + S &	49.30 &	31.78 &	46.89  & 56.58 & 36.43  & 52.06  & 44.35 & 24.73 & 41.46\\
BART + S + H &	49.96 &	32.72 &	47.66 & 58.26 &40.08  & 55.34  & 45.52 &25.71  &42.51 \\
BART + S + H + D (QFCL) &	\textbf{51.48} &	\textbf{34.16} & \textbf{49.08} & \textbf{60.09} & \textbf{43.22} & \textbf{57.54}  & \textbf{46.42} & \textbf{26.47} & \textbf{43.41} \\
\hline
\end{tabular}%
}
\caption{Experimental results on three medical question summarization datasets. $S$ denotes the contrastive learning on simple negative samples at the encoder end; $H$ denotes the contrastive learning on hard negative samples at the encoder end; $D$ denotes the decoder end’s contrastive learning. The top group lists the existing state-of-the-art results on three datasets, and the bottom group shows our ablation study on different components.}
\label{tab:result}
\end{table*}

\subsection{Training Details}
We utilize BART-large \cite{lewis-etal-2020-bart} in huggingface\footnote{\noindent huggingface.co/facebook/bart-large\label{bart_footnote}} as our pre-trained model. The learning rate of BART baseline is set to 3e-5 as the same with \citet{mrini-etal-2021-gradually}. For contrastive learning in QFCL, the learning rate is optimized to 1e-5. Betas of Adam optimizer is set to 0.9 and 0.999. Batch size is set to 16. The number of hard negative samples $n_h$ is set to 64. For Moco, the queue size $K$ is set to 4096, temperature $\tau$ is 0.07, and the momentum coefficient $m$ is 0.999. In Equation \ref{eq7}, $\alpha$ and $\beta$ are set to 1 and 0.5 respectively through grid search on MeqSum development set. 
Experiments were all performed on a single NVIDIA RTX 3090 GPU. The average runtimes of each epoch for MeqSum, iCliniq and HealthcareMagic are 4.2h, 0.6h and 0.1h respectively.

\subsection{Overall Performance}
We report our experimental results in Table \ref{tab:result}. Our model achieves new state-of-the-art results on all three datasets. Compared with the previous best results, we obtain an improvement of 0.99 ROUGE-L score on MeqSum, 8.44 on iCliniq, and 0.51 on HealthcareMagic, respectively.  

MTL+Data augmentation \cite{mrini-etal-2021-gradually} obtains the previous state-of-the-art results on iCliniq and HealthcareMagic, which utilizes the question entailment data to augment summarization data. In contrast, our method doesn't need other classification models or external data. The work of ProphetNet+QTR+QFR \cite{yadav-etal-2021-reinforcement} gets the previous best result on MeqSum, which presents a reinforcement learning-based framework with question-aware rewards. Comparing with this competitive model, our method obtains consistent better performance on all metrics, with 2.28 improvement on R1, 4.66 improvement on R2 and 0.89 improvement on RL. We did not compare the results of \cite{yadav-etal-2021-reinforcement} on the other two datasets, since their method requires manually labeled question focuses and question types.



\subsection{Ablation Study}
We perform ablation study to evaluate the impacts of different components employed in QFCL, and report the results in Table \ref{tab:result}. In particular, for Meqsum dataset, due to the small size which may cause the training unstable, we conducted five separate experiments and computed the average ROUGE score of these five checkpoints as the final result. Compared with the base BART model, we obtain an absolute improvement of 5.33 points on average. T-test is implemented on such five ROUGE scores and the p-value is less than 1e-2, validating that this improvement is significant. On Cliniq the absolute improvement is 12.85 points and on HealthcareMagic 3.81 points. In comparison to BART, the relative improvements of our model are $12.2\%$, $28.7\%$ and $9.6\%$ on Meqsum, Cliniq and HealthcareMagic respectively. 

The results demonstrate that each component of our model is helpful. On MeqSum, there is an increase of 3.15 points for BART+S compared to the baseline, indicating that the contrastive learning on simple negative samples largely improves model performance. It shows an continuous increase of 0.77 points for BART+S+H, and the highest ROUGE-L score is obtained when three parts are all implemented in our model. It suggests that each component in QFCL contributes positively, and metrics like ROUGE evaluating the similarity between whole sentences benefit from our contrastive learning strategy.

\subsection{Human Evaluation}
To quantitatively assess the results, we compare our method with the baseline BART  through human judgement. We randomly selected 50 samples from each of three datasets, and hired 3 graduate students to categorize each generated summary into one of the following categories: 'Incorrect', 'Acceptable', and 'Perfect'. We compute the average number of each category, and report the result in Table \ref{tab:humaneval}. The average Spearman correlation coefficient between three annotators is 0.68, which guarantees a high quality of our annotation data. The evaluation results show that our model generates a higher proportion of perfect samples and a lower proportion of incorrect ones, by enhancing the model's ability of capturing sentence semantics and question focuses.

\begin{table}[h]
\small
\setlength\tabcolsep{2pt}
\centering
\begin{tabular}{l|ccc|ccc|ccc}
\hline
\multirow{2}{*}{Model} &\multicolumn{3}{c|}{MeqSum}  &\multicolumn{3}{c|}{iCliniq}  &\multicolumn{3}{c}{HealthCareMagic}\\ 
\cline{2-10} 
& I  & A  & P  & I & A & P & I  & A  & P \\ 
\hline
\hline
BART &   28.7 & 17.3 & 4.0 & 12.3 & 17.0 & 20.7 & 20.7  & 20.3 & 9.0 \\ 
QFCL	&   12.0 & 18.0 & 20.0 & 6.3 & 17.7 & 26.0 & 5.7  & 16.3 & 28.0 \\
\hline
\end{tabular}%
\caption{Human evaluation of the summaries generated by BART and QFCL respectively. The metric \textbf{I} means the number of incorrect samples, \textbf{A} means acceptable, \textbf{P} means perfect. }
\label{tab:humaneval}
\end{table}

\subsection{Case Study}
To clearly show the output question summary,
we list two samples to compare our model with BART in Table \ref{table:case}. In Case 1, BART captures the question focus "\textit{Ampicillin}" but misses "\textit{drink alcohol}", and in Case 2 it misses the question focus "\textit{breast milk}". In contrast, our model successfully extract multiple question focuses from the lengthy CHQ, and generate summaries which more conform to the meaning of original questions.  

\begin{table}[ht]
\resizebox{\linewidth}{!}{%
\begin{tabular}{c|l}
\hline
\multicolumn{2}{c}{\textbf{Case1}}\\
\hline
\multirow{3}{*}{\textbf{CHQ}} & MESSAGE: Is it okay to drink \textbf{alcohol} in \\
& moderation when taking \textbf{Ampicillin}.  I was\\
&told it negates any medical effect of the drug\\
\hline
\textbf{FAQ}&
Can I drink \textbf{alcohol} while taking \textbf{Amoxicillin}?\\
\hline
\hline
\textbf{BART}&
What are the side effects of \textbf{Ampicillin}? \\
\hline
\textbf{QFCL}& 
Is it okay to drink \textbf{alcohol} with \textbf{Ampicillin}? \\
\hline
\multicolumn{2}{c}{\textbf{Case2}}\\
\hline
\multirow{9}{*}{\textbf{CHQ}} & Hi..... I have 3 month old baby girl...... I don t \\
&have \textbf{breast milk} from the beginning due to\\
& some reason. I can not give formula milk to \\
& baby...... So right now i m giving buffelo milk\\
& ........ What else i should give her for better\\
&nourishment????? ....... She has constipation\\
& problem may be due to milk but i cant give her\\
&breastmilk or formula ....... How to overcome \\
&it?????......... Please help me\\
\hline
\multirow{2}{*}{\textbf{FAQ}}&
Suggest ways to feed newborn other than \\
& \textbf{breast milk}\\
\hline
\hline
\textbf{BART}& 
Suggest treatment for constipation in a child \\
\hline
\multirow{2}{*}{\textbf{QFCL}}& 
Suggest better nourishment for baby other\\
& than \textbf{breast milk} \\
\hline
\end{tabular}
}
\caption{Examples of generated question summaries by BART and our QFCL model. The question focuses are highlighted.}
\label{table:case}
\end{table}

\begin{figure}[!!ht]
\centering
\includegraphics[width=\linewidth]{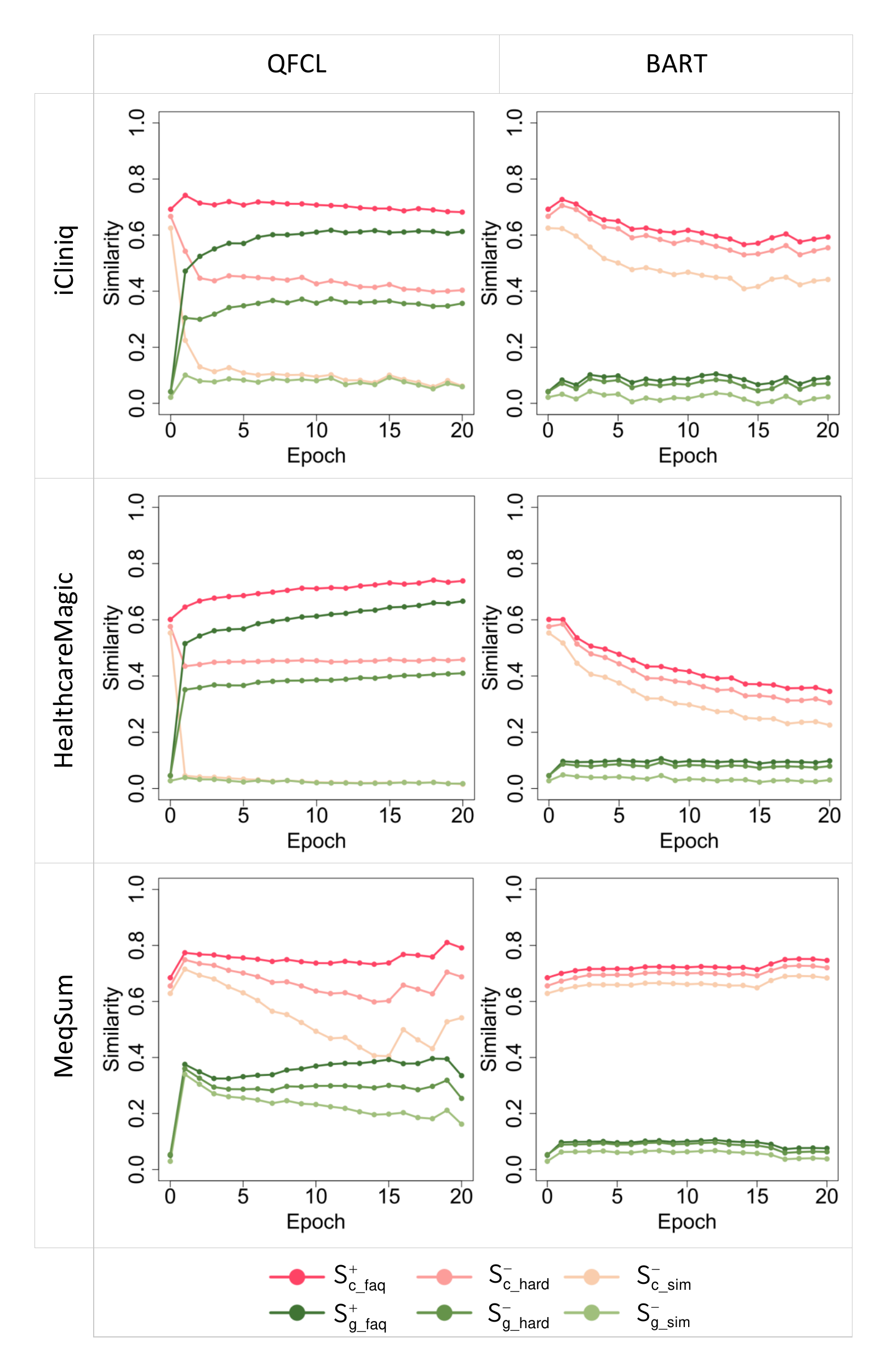} 
\caption{Correlation between sentence representation similarities and epoch numbers on dev set. The red lines are about the anchor CHQ. $s_{c\_faq}^{+}$ is the average cosine similarity between CHQ and related FAQ, $s_{c\_sim}^{-}$ is between CHQ and simple negative samples (other FAQs), $s_{c\_hard}^{-}$ is between CHQ and hard negative samples. The green lines are about the anchor of generated summary. $s_{g\_faq}^{+}$ is the average cosine similarity between the generated summary and FAQ, $s_{g\_sim}^{-}$ is between generated summary and simple negatives, $s_{g\_hard}^{-}$ is between generated summary and hard negatives. The epoch number equaling 0 denotes the initial pre-trained model.}
\label{similarity_and_epoch}
\end{figure}

\section{Model Analysis}
\subsection{Correlation of Sentence Representations}
Since the auxiliary structures are discarded at the inference stage, we make further analysis to check that whether the retained model has the ability to distinguish different sentence-level semantics when facing unknown data. We train QFCL and BART on the training set for 20 epochs and save each checkpoint, and evaluate these checkpoints on the development set. 

Four types of sentence representations are extracted from these checkpoints: CHQ’s representation $\mathcal{R}_c$, FAQ’s representation $\mathcal{R}_f$, hard negatives’ representation $\mathcal{R}_h$, and the generated summary’s representation at decoder end $\mathcal{R}_g$. Then we calculate the cosine similarity between them, and draw the relationship between these similarity scores and the epoch numbers, as shown in Figure \ref{similarity_and_epoch}. 


Regarding the anchor CHQ in the curve of iCliniq, \begin{small}$s_{c\_faq}^{+}$\end{small}, \begin{small}$s_{c\_sim}^{-}$\end{small} and \begin{small}$s_{c\_hard}^{-}$\end{small} are very close to each other at epoch 0, suggesting that the initial encoder lacks the ability to capture different semantics. With the increase of training steps, \begin{small}$s_{c\_faq}^{+}$\end{small} changes smoothly, while \begin{small}$s_{c\_sim}^{-}$\end{small} decreases sharply to near zero and \begin{small}$s_{c\_hard}^{-}$\end{small} decreases gradually and converges at a middle level between \begin{small}$s_{c\_faq}^{+}$\end{small} and \begin{small}$s_{c\_sim}^{-}$\end{small}. 
This suggests that, powered by contrastive learning, our model has learned to distinguish sentences of different meanings at the encoder end.  

With the generated summary as another anchor, we find out that \begin{small}$s_{g\_faq}^{+}$\end{small}, \begin{small}$s_{g\_sim}^{-}$\end{small}, \begin{small}$s_{g\_hard}^{-}$\end{small} are all near to 0 initially, which depict that the decoder is also weak in representing sentence-level semantics. After training, \begin{small}$s_{g\_faq}^{+}$\end{small} increases significantly, \begin{small}$s_{g\_hard}^{-}$\end{small} converges between \begin{small}$s_{g\_faq}^{+}$\end{small} and \begin{small}$s_{g\_sim}^{-}$\end{small}, and \begin{small}$s_{g\_sim}^{-}$\end{small} keeps very low all the time. It suggests that the decoder has strengthened its power to distinguish different semantics as the same to the encoder end.

Another chart is drawn to show this relationship for BART baseline in Figure \ref{similarity_and_epoch}. The similarities between the anchor and the positive samples, negative samples are very close, and never improve significantly with the progress of training. This situation suggesting that the BART baseline has a relatively weaker performance to distinguish the sentences of different meanings at both encoder and decode, since it only focuses on the prediction of next tokens. 

We also draw this correlation curve on MeqSum and HealthcareMagic. The curve of HealthcareMagic is similar to iCliniq. On MeqSum, our model can still distinguish sentences with different semantics better than the baseline, but the signal is not as significant as iCliniq or HealthcareMagic due to the limited size of training set.

\begin{table}
\centering
\resizebox{\linewidth}{!}{%
\begin{tabular}{ccccccc}
\hline
Model   &    C1   & C2   & C3       & C4 &C5 & Mean     \\ 
\hline
\hline
BART     & 33.37 & 40.76 & 39.78 & 35.34 & 36.21 & 37.09   \\ 
QFCL        & 47.41 & 42.24 & 45.20  & 45.20 &47.17 & \textbf{45.44}  \\ \hline
\end{tabular}%
}
\caption{Accuracy of question focuses in generated summaries. C1-C5 means 5 different checkpoints trained by each model.}
\label{tab:question_focus}
\end{table}

\subsection{Capturing Question Focus}
To study whether our model pays more attention to the question focus, we evaluate the accuracy of question focuses in generated summaries. We use the sequence labeling model trained by \citet{yadav-etal-2021-reinforcement} to predict question focuses on the MeqSum dataset, 
and regard the 812 predicted question focuses in test set as the gold-standard. For QFCL and BART, we train five checkpoints and generate summaries on these checkpoints, and compute the accuracy of question focuses on test set.
As shown in Table \ref{tab:question_focus}, the average accuracy is 37.09\% for BART and 45.44\% for QFCL. Our model exceeds the baseline by 8.35 points for question focus generation. P-value of t-test on these two sets of results is 1.04e-3, indicating that this improvement is statistically significant.

\section{Conclusion}
In this paper, we introduce a novel question focus-based contrastive learning framework QFCL for medical question summarization. In the proposed model, we adopt a "double anchor" strategy, by considering both the input question CHQ and the generated summary as comparing anchors. And we exploit a "hard negatives generator" to generate hard negative samples based on the question focus. 
Our model significantly improves the performance on three medical question summarization datasets, and achieves new state-of-the-art results. 
In the future, we would like to find a more effective way  to do question focus recognition.

\section*{Acknowledgement}
This work is supported by the National Hi-Tech RD Program of China (No.2020AAA0106600), the  National Natural  Science  Foundation  of  China  (62076008) and the Key Project of Natural Science Foundation of China (61936012).

\appendix
\section{Ethical Consideration}
\label{sec:appendix}
The datasets used in our work are all publicly available. We used BART as our basic model which follows the apache-2.0 license. The datasets and the method should only be used for research purposes, not in the commercial field.

The personal information in the datasets has been hidden through preprocessing. For example, the name of the consumer was converted to an placeholder [name] and the address was converted to [location], as shown in Table \ref{fig1}.

As the current models could not guarantee to generate summaries fully conforms to the intention of the consumers, the method in our paper can only be used as an auxiliary tool to avoid further misleading suggestions.

\bibliography{custom}

\begin{thebibliography}{29}
\expandafter\ifx\csname natexlab\endcsname\relax\def\natexlab#1{#1}\fi

\bibitem[{Akbik et~al.(2018)Akbik, Blythe, and
  Vollgraf}]{akbik-etal-2018-contextual}
Alan Akbik, Duncan Blythe, and Roland Vollgraf. 2018.
\newblock \href {https://aclanthology.org/C18-1139} {Contextual string
  embeddings for sequence labeling}.
\newblock In \emph{Proceedings of the 27th International Conference on
  Computational Linguistics}, pages 1638--1649, Santa Fe, New Mexico, USA.
  Association for Computational Linguistics.

\bibitem[{Ben~Abacha and
  Demner-Fushman(2019)}]{ben-abacha-demner-fushman-2019-summarization}
Asma Ben~Abacha and Dina Demner-Fushman. 2019.
\newblock \href {https://doi.org/10.18653/v1/P19-1215} {On the summarization of
  consumer health questions}.
\newblock In \emph{Proceedings of the 57th Annual Meeting of the Association
  for Computational Linguistics}, pages 2228--2234, Florence, Italy.
  Association for Computational Linguistics.

\bibitem[{Ben~Abacha et~al.(2021)Ben~Abacha, Mrabet, Zhang, Shivade, Langlotz,
  and Demner-Fushman}]{ben-abacha-etal-2021-overview}
Asma Ben~Abacha, Yassine Mrabet, Yuhao Zhang, Chaitanya Shivade, Curtis
  Langlotz, and Dina Demner-Fushman. 2021.
\newblock \href {https://doi.org/10.18653/v1/2021.bionlp-1.8} {Overview of the
  {MEDIQA} 2021 shared task on summarization in the medical domain}.
\newblock In \emph{Proceedings of the 20th Workshop on Biomedical Language
  Processing}, pages 74--85, Online. Association for Computational Linguistics.

\bibitem[{Cao and Wang(2021)}]{cao-wang-2021-cliff}
Shuyang Cao and Lu~Wang. 2021.
\newblock \href {https://aclanthology.org/2021.emnlp-main.532} {{CLIFF}:
  Contrastive learning for improving faithfulness and factuality in abstractive
  summarization}.
\newblock In \emph{Proceedings of the 2021 Conference on Empirical Methods in
  Natural Language Processing}, pages 6633--6649, Online and Punta Cana,
  Dominican Republic. Association for Computational Linguistics.

\bibitem[{Chen et~al.(2020)Chen, Kornblith, Norouzi, and
  Hinton}]{chen2020simple}
Ting Chen, Simon Kornblith, Mohammad Norouzi, and Geoffrey Hinton. 2020.
\newblock A simple framework for contrastive learning of visual
  representations.
\newblock In \emph{International conference on machine learning}, pages
  1597--1607. PMLR.

\bibitem[{Chi et~al.(2021)Chi, Dong, Wei, Yang, Singhal, Wang, Song, Mao,
  Huang, and Zhou}]{chi-etal-2021-infoxlm}
Zewen Chi, Li~Dong, Furu Wei, Nan Yang, Saksham Singhal, Wenhui Wang, Xia Song,
  Xian-Ling Mao, Heyan Huang, and Ming Zhou. 2021.
\newblock \href {https://doi.org/10.18653/v1/2021.naacl-main.280} {{I}nfo{XLM}:
  An information-theoretic framework for cross-lingual language model
  pre-training}.
\newblock In \emph{Proceedings of the 2021 Conference of the North American
  Chapter of the Association for Computational Linguistics: Human Language
  Technologies}, pages 3576--3588, Online. Association for Computational
  Linguistics.

\bibitem[{Greensmith et~al.(2004)Greensmith, Bartlett, and
  Baxter}]{greensmith2004variance}
Evan Greensmith, Peter~L Bartlett, and Jonathan Baxter. 2004.
\newblock Variance reduction techniques for gradient estimates in reinforcement
  learning.
\newblock \emph{Journal of Machine Learning Research}, 5(9).

\bibitem[{He et~al.(2020)He, Fan, Wu, Xie, and Girshick}]{he2020momentum}
Kaiming He, Haoqi Fan, Yuxin Wu, Saining Xie, and Ross Girshick. 2020.
\newblock Momentum contrast for unsupervised visual representation learning.
\newblock In \emph{Proceedings of the IEEE/CVF Conference on Computer Vision
  and Pattern Recognition}, pages 9729--9738.

\bibitem[{He et~al.(2021)He, Chen, and Huang}]{he-etal-2021-damo}
Yifan He, Mosha Chen, and Songfang Huang. 2021.
\newblock \href {https://doi.org/10.18653/v1/2021.bionlp-1.12} {damo{\_}nlp at
  {MEDIQA} 2021: Knowledge-based preprocessing and coverage-oriented reranking
  for medical question summarization}.
\newblock In \emph{Proceedings of the 20th Workshop on Biomedical Language
  Processing}, pages 112--118, Online. Association for Computational
  Linguistics.

\bibitem[{Henaff(2020)}]{henaff2020data}
Olivier Henaff. 2020.
\newblock Data-efficient image recognition with contrastive predictive coding.
\newblock In \emph{International Conference on Machine Learning}, pages
  4182--4192. PMLR.

\bibitem[{Kalantidis et~al.(2020)Kalantidis, Sariyildiz, Pion, Weinzaepfel, and
  Larlus}]{NEURIPS2020_f7cade80}
Yannis Kalantidis, Mert~Bulent Sariyildiz, Noe Pion, Philippe Weinzaepfel, and
  Diane Larlus. 2020.
\newblock \href
  {https://proceedings.neurips.cc/paper/2020/file/f7cade80b7cc92b991cf4d2806d6bd78-Paper.pdf}
  {Hard negative mixing for contrastive learning}.
\newblock In \emph{Advances in Neural Information Processing Systems},
  volume~33, pages 21798--21809. Curran Associates, Inc.

\bibitem[{Lewis et~al.(2020)Lewis, Liu, Goyal, Ghazvininejad, Mohamed, Levy,
  Stoyanov, and Zettlemoyer}]{lewis-etal-2020-bart}
Mike Lewis, Yinhan Liu, Naman Goyal, Marjan Ghazvininejad, Abdelrahman Mohamed,
  Omer Levy, Veselin Stoyanov, and Luke Zettlemoyer. 2020.
\newblock \href {https://doi.org/10.18653/v1/2020.acl-main.703} {{BART}:
  Denoising sequence-to-sequence pre-training for natural language generation,
  translation, and comprehension}.
\newblock In \emph{Proceedings of the 58th Annual Meeting of the Association
  for Computational Linguistics}, pages 7871--7880, Online. Association for
  Computational Linguistics.

\bibitem[{Li et~al.(2020)Li, Zhu, Zhang, Zong, and He}]{li2020keywords}
Haoran Li, Junnan Zhu, Jiajun Zhang, Chengqing Zong, and Xiaodong He. 2020.
\newblock Keywords-guided abstractive sentence summarization.
\newblock In \emph{Proceedings of the AAAI Conference on Artificial
  Intelligence}, volume~34, pages 8196--8203.

\bibitem[{Li et~al.(2019)Li, Lei, Qin, and Wang}]{li-etal-2019-deep}
Siyao Li, Deren Lei, Pengda Qin, and William~Yang Wang. 2019.
\newblock \href {https://doi.org/10.18653/v1/D19-1623} {Deep reinforcement
  learning with distributional semantic rewards for abstractive summarization}.
\newblock In \emph{Proceedings of the 2019 Conference on Empirical Methods in
  Natural Language Processing and the 9th International Joint Conference on
  Natural Language Processing (EMNLP-IJCNLP)}, pages 6038--6044, Hong Kong,
  China. Association for Computational Linguistics.

\bibitem[{Lin(2004)}]{lin-2004-rouge}
Chin-Yew Lin. 2004.
\newblock \href {https://aclanthology.org/W04-1013} {{ROUGE}: A package for
  automatic evaluation of summaries}.
\newblock In \emph{Text Summarization Branches Out}, pages 74--81, Barcelona,
  Spain. Association for Computational Linguistics.

\bibitem[{Liu and Liu(2021)}]{liu-liu-2021-simcls}
Yixin Liu and Pengfei Liu. 2021.
\newblock \href {https://doi.org/10.18653/v1/2021.acl-short.135} {{S}im{CLS}: A
  simple framework for contrastive learning of abstractive summarization}.
\newblock In \emph{Proceedings of the 59th Annual Meeting of the Association
  for Computational Linguistics and the 11th International Joint Conference on
  Natural Language Processing (Volume 2: Short Papers)}, pages 1065--1072,
  Online. Association for Computational Linguistics.

\bibitem[{Misra and van~der Maaten(2020)}]{9156540}
Ishan Misra and Laurens van~der Maaten. 2020.
\newblock \href {https://doi.org/10.1109/CVPR42600.2020.00674} {Self-supervised
  learning of pretext-invariant representations}.
\newblock In \emph{2020 IEEE/CVF Conference on Computer Vision and Pattern
  Recognition (CVPR)}, pages 6706--6716.

\bibitem[{Mrini et~al.(2021{\natexlab{a}})Mrini, Dernoncourt, Chang, Farcas,
  and Nakashole}]{mrini-etal-2021-joint}
Khalil Mrini, Franck Dernoncourt, Walter Chang, Emilia Farcas, and Ndapa
  Nakashole. 2021{\natexlab{a}}.
\newblock \href {https://doi.org/10.18653/v1/2021.nlpmc-1.8} {Joint
  summarization-entailment optimization for consumer health question
  understanding}.
\newblock In \emph{Proceedings of the Second Workshop on Natural Language
  Processing for Medical Conversations}, pages 58--65, Online. Association for
  Computational Linguistics.

\bibitem[{Mrini et~al.(2021{\natexlab{b}})Mrini, Dernoncourt, Yoon, Bui, Chang,
  Farcas, and Nakashole}]{mrini-etal-2021-gradually}
Khalil Mrini, Franck Dernoncourt, Seunghyun Yoon, Trung Bui, Walter Chang,
  Emilia Farcas, and Ndapa Nakashole. 2021{\natexlab{b}}.
\newblock \href {https://doi.org/10.18653/v1/2021.acl-long.119} {A gradually
  soft multi-task and data-augmented approach to medical question
  understanding}.
\newblock In \emph{Proceedings of the 59th Annual Meeting of the Association
  for Computational Linguistics and the 11th International Joint Conference on
  Natural Language Processing (Volume 1: Long Papers)}, pages 1505--1515,
  Online. Association for Computational Linguistics.

\bibitem[{Mrini et~al.(2021{\natexlab{c}})Mrini, Dernoncourt, Yoon, Bui, Chang,
  Farcas, and Nakashole}]{mrini-etal-2021-ucsd}
Khalil Mrini, Franck Dernoncourt, Seunghyun Yoon, Trung Bui, Walter Chang,
  Emilia Farcas, and Ndapa Nakashole. 2021{\natexlab{c}}.
\newblock \href {https://doi.org/10.18653/v1/2021.bionlp-1.28} {{UCSD}-adobe at
  {MEDIQA} 2021: Transfer learning and answer sentence selection for medical
  summarization}.
\newblock In \emph{Proceedings of the 20th Workshop on Biomedical Language
  Processing}, pages 257--262, Online. Association for Computational
  Linguistics.

\bibitem[{Nallapati et~al.(2016)Nallapati, Zhou, dos Santos, Gul{\c{c}}ehre,
  and Xiang}]{nallapati-etal-2016-abstractive}
Ramesh Nallapati, Bowen Zhou, Cicero dos Santos, {\c{C}}a{\u{g}}lar
  Gul{\c{c}}ehre, and Bing Xiang. 2016.
\newblock \href {https://doi.org/10.18653/v1/K16-1028} {Abstractive text
  summarization using sequence-to-sequence {RNN}s and beyond}.
\newblock In \emph{Proceedings of The 20th {SIGNLL} Conference on Computational
  Natural Language Learning}, pages 280--290, Berlin, Germany. Association for
  Computational Linguistics.

\bibitem[{Pan et~al.(2021)Pan, Wang, Wu, and Li}]{pan-etal-2021-contrastive}
Xiao Pan, Mingxuan Wang, Liwei Wu, and Lei Li. 2021.
\newblock \href {https://doi.org/10.18653/v1/2021.acl-long.21} {Contrastive
  learning for many-to-many multilingual neural machine translation}.
\newblock In \emph{Proceedings of the 59th Annual Meeting of the Association
  for Computational Linguistics and the 11th International Joint Conference on
  Natural Language Processing (Volume 1: Long Papers)}, pages 244--258, Online.
  Association for Computational Linguistics.

\bibitem[{Paulus et~al.(2018)Paulus, Xiong, and Socher}]{paulus2018a}
Romain Paulus, Caiming Xiong, and Richard Socher. 2018.
\newblock \href {https://openreview.net/forum?id=HkAClQgA-} {A deep reinforced
  model for abstractive summarization}.
\newblock In \emph{International Conference on Learning Representations}.

\bibitem[{S{\"a}nger et~al.(2021)S{\"a}nger, Weber, and
  Leser}]{sanger-etal-2021-wbi}
Mario S{\"a}nger, Leon Weber, and Ulf Leser. 2021.
\newblock \href {https://doi.org/10.18653/v1/2021.bionlp-1.9} {{WBI} at
  {MEDIQA} 2021: Summarizing consumer health questions with generative
  transformers}.
\newblock In \emph{Proceedings of the 20th Workshop on Biomedical Language
  Processing}, pages 86--95, Online. Association for Computational Linguistics.

\bibitem[{Yadav et~al.(2021{\natexlab{a}})Yadav, Gupta, Ben~Abacha, and
  Demner-Fushman}]{yadav-etal-2021-reinforcement}
Shweta Yadav, Deepak Gupta, Asma Ben~Abacha, and Dina Demner-Fushman.
  2021{\natexlab{a}}.
\newblock \href {https://doi.org/10.18653/v1/2021.acl-short.33} {Reinforcement
  learning for abstractive question summarization with question-aware semantic
  rewards}.
\newblock In \emph{Proceedings of the 59th Annual Meeting of the Association
  for Computational Linguistics and the 11th International Joint Conference on
  Natural Language Processing (Volume 2: Short Papers)}, pages 249--255,
  Online. Association for Computational Linguistics.

\bibitem[{Yadav et~al.(2021{\natexlab{b}})Yadav, Sarrouti, and
  Gupta}]{yadav-etal-2021-nlm}
Shweta Yadav, Mourad Sarrouti, and Deepak Gupta. 2021{\natexlab{b}}.
\newblock \href {https://doi.org/10.18653/v1/2021.bionlp-1.34} {{NLM} at
  {MEDIQA} 2021: Transfer learning-based approaches for consumer question and
  multi-answer summarization}.
\newblock In \emph{Proceedings of the 20th Workshop on Biomedical Language
  Processing}, pages 291--301, Online. Association for Computational
  Linguistics.

\bibitem[{Yang et~al.(2021)Yang, Wei, Jiao, Jiang, and
  Yang}]{yang-etal-2021-xmoco}
Nan Yang, Furu Wei, Binxing Jiao, Daxing Jiang, and Linjun Yang. 2021.
\newblock \href {https://doi.org/10.18653/v1/2021.acl-long.477} {x{M}o{C}o:
  Cross momentum contrastive learning for open-domain question answering}.
\newblock In \emph{Proceedings of the 59th Annual Meeting of the Association
  for Computational Linguistics and the 11th International Joint Conference on
  Natural Language Processing (Volume 1: Long Papers)}, pages 6120--6129,
  Online. Association for Computational Linguistics.

\bibitem[{Zeng et~al.(2020)Zeng, Yang, Ju, Yang, Wang, Zhang, Zhou, Zeng, Dong,
  Zhang, Fang, Zhu, Chen, and Xie}]{zeng-etal-2020-meddialog}
Guangtao Zeng, Wenmian Yang, Zeqian Ju, Yue Yang, Sicheng Wang, Ruisi Zhang,
  Meng Zhou, Jiaqi Zeng, Xiangyu Dong, Ruoyu Zhang, Hongchao Fang, Penghui Zhu,
  Shu Chen, and Pengtao Xie. 2020.
\newblock \href {https://doi.org/10.18653/v1/2020.emnlp-main.743}
  {{M}ed{D}ialog: Large-scale medical dialogue datasets}.
\newblock In \emph{Proceedings of the 2020 Conference on Empirical Methods in
  Natural Language Processing (EMNLP)}, pages 9241--9250, Online. Association
  for Computational Linguistics.

\bibitem[{Zhang et~al.(2020)Zhang, Zhao, Saleh, and Liu}]{zhang2020pegasus}
Jingqing Zhang, Yao Zhao, Mohammad Saleh, and Peter Liu. 2020.
\newblock Pegasus: Pre-training with extracted gap-sentences for abstractive
  summarization.
\newblock In \emph{International Conference on Machine Learning}, pages
  11328--11339. PMLR.

\end{thebibliography}
\bibliographystyle{acl_natbib}

\end{document}